\documentclass[10pt,twocolumn,letterpaper]{article}

\usepackage{iccv}              %
\PassOptionsToPackage{capitalise,noabbrev,nameinlink}{cleveref}

\usepackage[most]{tcolorbox}
\usepackage{graphicx}
\usepackage{multirow}

\definecolor{iccvblue}{rgb}{0.21,0.49,0.74}
\usepackage[pagebackref,breaklinks,colorlinks,allcolors=iccvblue]{hyperref}

\usepackage{wrapfig}

\def\paperID{4303} %
\def\confName{ICCV}
\def\confYear{2025}

\title{Visual Test-time Scaling for GUI Agent Grounding}

\author{Tiange Luo$^{1,2}$ \qquad Lajanugen Logeswaran$^{2,\dagger}$ \qquad Justin Johnson$^{1,\dagger}$  \qquad 
Honglak Lee$^{1,2,\dagger}$  \\
University of Michigan$^1$ \qquad LG AI Research$^2$ \qquad $^\dagger$ equal advising
}

\begin{document}
\maketitle

\begin{abstract}
We introduce RegionFocus, a visual test-time scaling approach for Vision Language Model Agents. Understanding webpages is challenging due to the visual complexity of GUI images and the large number of interface elements, making accurate action selection difficult. Our approach dynamically zooms in on relevant regions, reducing background clutter and improving grounding accuracy. To support this process, we propose an image-as-map mechanism that visualizes key landmarks at each step, providing a transparent action record and enables the agent to effectively choose among action candidates. 
Even with a simple region selection strategy, we observe significant performance gains of 28+\% on Screenspot-pro and 24+\% on WebVoyager benchmarks on top of two state-of-the-art open vision language model agents, UI-TARS and Qwen2.5-VL, highlighting the effectiveness of visual test-time scaling in interactive settings.
We achieve a new state-of-the-art grounding performance of 61.6\% on the ScreenSpot-Pro benchmark by applying RegionFocus to a Qwen2.5-VL-72B model.
Our code will be released publicly at \href{https://github.com/tiangeluo/RegionFocus}{https://github.com/tiangeluo/RegionFocus}. 

\end{abstract}
    
\section{Introduction}
\label{sec:intro}

Graphical user interface (GUI) agents have become increasingly pivotal in modern computing, powering applications ranging from automated web browsing to intuitive operating system navigation \cite{anderson2018vision,liu2024visualagentbench}. With the proliferation of large-scale vision-language models (VLMs), researchers have sought to harness both textual and visual information to build more capable interactive systems \cite{anthropic2024computeruse}. While many existing frameworks rely heavily on text-based reasoning~\cite{zheng2024gpt, yang2024agentoccamsimplestrongbaseline} or simple visual grounding~\cite{lu2024omniparserpurevisionbased, gou2024navigating}, real-world GUIs often contain a substantial number of irrelevant elements—such as menu bars, ads, and extraneous buttons—that can overwhelm purely textual or naive visual approaches. This mismatch between text-heavy inference and the visual complexity of GUIs leads to frequent errors (e.g., clicking the wrong button or navigating to an unintended section). Since these tasks are typically high-level, such low-level mistakes accumulate and ultimately result in higher failure rates and poorer overall performance.

Recent research on GUI agents typically falls into two main categories: those relying on textual cues for planning and reasoning, and those incorporating visual information through VLMs. Text-based approaches often generate text labels or bounding boxes for each visual element to guide agent actions~\cite{zheng2024gpt, lu2024omniparserpurevisionbased}. However, they can struggle with visually entangled tasks where textual descriptions are ambiguous, incomplete, or fail to capture crucial visual features (e.g., floating windows), even when using accessibility trees. On the other hand, vision-based pipelines~\cite{gou2024navigating, qin2025ui} often rely heavily on the VLM’s ability to ground visual elements. We observe that many errors arise from inadvertently clicking empty or incorrect interface components, underscoring the limitations of existing single-inference visual grounding methods using VLMs. Once an error occurs, there is no feedback loop to correct it, causing mistakes to compound throughout the process.

\begin{figure}
    \centering
    \includegraphics[width=1\linewidth,trim={1em 1em 1em 0},clip]{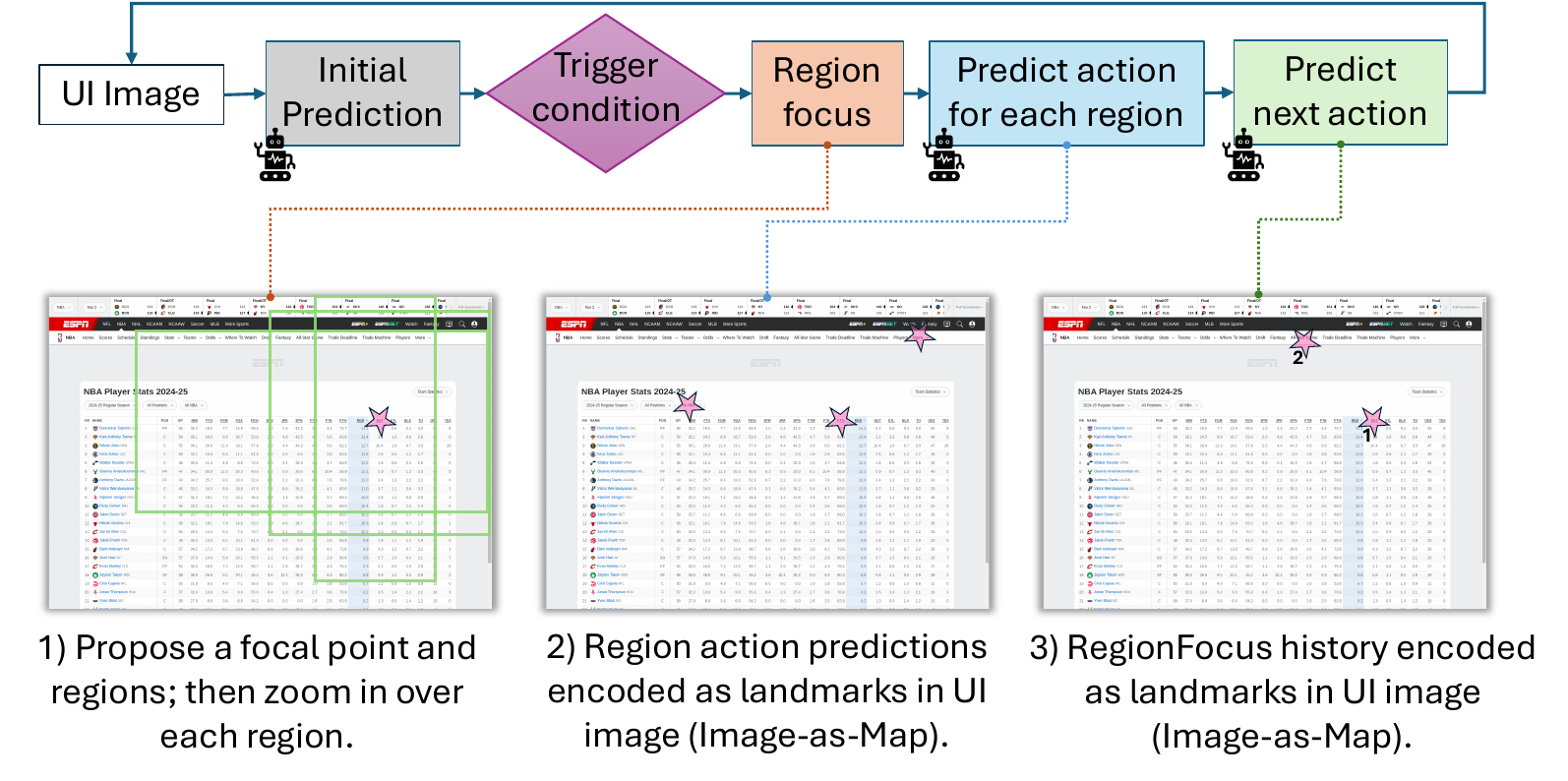}
    \vspace{-0.25in}
    \caption{\textbf{Overview}. When GUI agents encounter execution errors, we instruct the model to focus on a specific point of interest and extract multiple sub-regions around this focal point (1). The agent then independently generates candidate actions for each sub-region. Actions that interact with a specific coordinate are marked with pink-star landmarks (e.g., “click”) to visually indicate the relevant location (2). We retain the pink-star landmarks to track interaction history of RegionFocus for diverse exploration (3).}
    \label{fig:teaser}
\end{figure}

Motivated by these shortcomings, we propose a \emph{visual test-time scaling} framework, RegionFocus, designed to narrow the GUI model's attention to salient interface regions when execution errors occur or other conditions are triggered (e.g., VLM self-judgment). Specifically, as illustrated in Figure~\ref{fig:teaser}(1), we leverage the VLM’s capability to identify points of interest and combine this with bounding-box proposals generated either from fixed-ratio masks or segmentation models such as SAM \cite{kirillov2023segment}. For each sub-region, the agent independently predicts actions based solely on the local context (Figure~\ref{fig:teaser}(2)), subsequently aggregating the top candidate actions to form a refined, single-step response. Furthermore, interactions with web or OS interfaces allow our method to zoom into targeted areas, enhancing the resolution of selected regions for more careful examination.
RegionFocus works as a modular plug-in for GUI agents without affecting the original workflow.

In order to keep track of regions visited by RegionFocus, we introduce an \emph{image-as-map} mechanism to record temporal information.
In this approach, elements previously considered by the agent are annotated in the UI screenshot with visual landmarks (e.g., the pink-stars in Figure~\ref{fig:teaser} (3)).
These landmarks prevent the agent from revisiting regions it has already examined and guide the agent towards unexplored areas.
These markers do not interfere with the model’s regular inference process which leverages unaltered webpage screenshots, and are only used in the RegionFocus component. 
Once the agent navigates to a new page, all landmarks are refreshed to have a new RegionFocus history.

In addition to representing RegionFocus history, we also leverage image-as-map in the action aggregation process (Figure~\ref{fig:teaser} (2)), where candidate actions are represented using landmarks for selecting an optimal action.
We find image-as-map to be highly effective in representing both temporal information (e.g., previously visited regions) and spatial information (e.g., multiple action candidates) compared to alternative representations such as element coordinates represented in the form of text alone.
This is particularly crucial for distinguishing between screen elements in close proximity, which is challenging to reason about based on a text representation of the coordinates alone.

With our proposed visual test-time scaling framework, we help existing models—such as UI-TARS~\cite{qin2025ui} and QWen2.5-VL~\cite{Qwen2.5-VL} —achieve better performance in both web-based and desktop interfaces. In particular, we demonstrate substantial performance gains on benchmarks including \textit{ScreenSpot-Pro} \cite{li2024screenspotpro} for OS-level GUI navigation, as well as \textit{WebVoyager} \cite{he2024webvoyager} for browser automation. Through our experiments, we show that even a simple fixed-ratio bounding-box generation approach yields pronounced improvements over baseline systems, underscoring the efficacy of focusing the model's attention on visually relevant regions. Our empirical studies further indicate that image-as-map consistently outperforms text-based representations for VLM agents. Overall, our findings highlight the value of visual test-time scaling as a simple yet powerful extension to existing VLM-based GUI agents.

\begin{figure*}[t]
    \centering
    \includegraphics[width=1\linewidth,trim={2em 1em 5em 0em},clip]{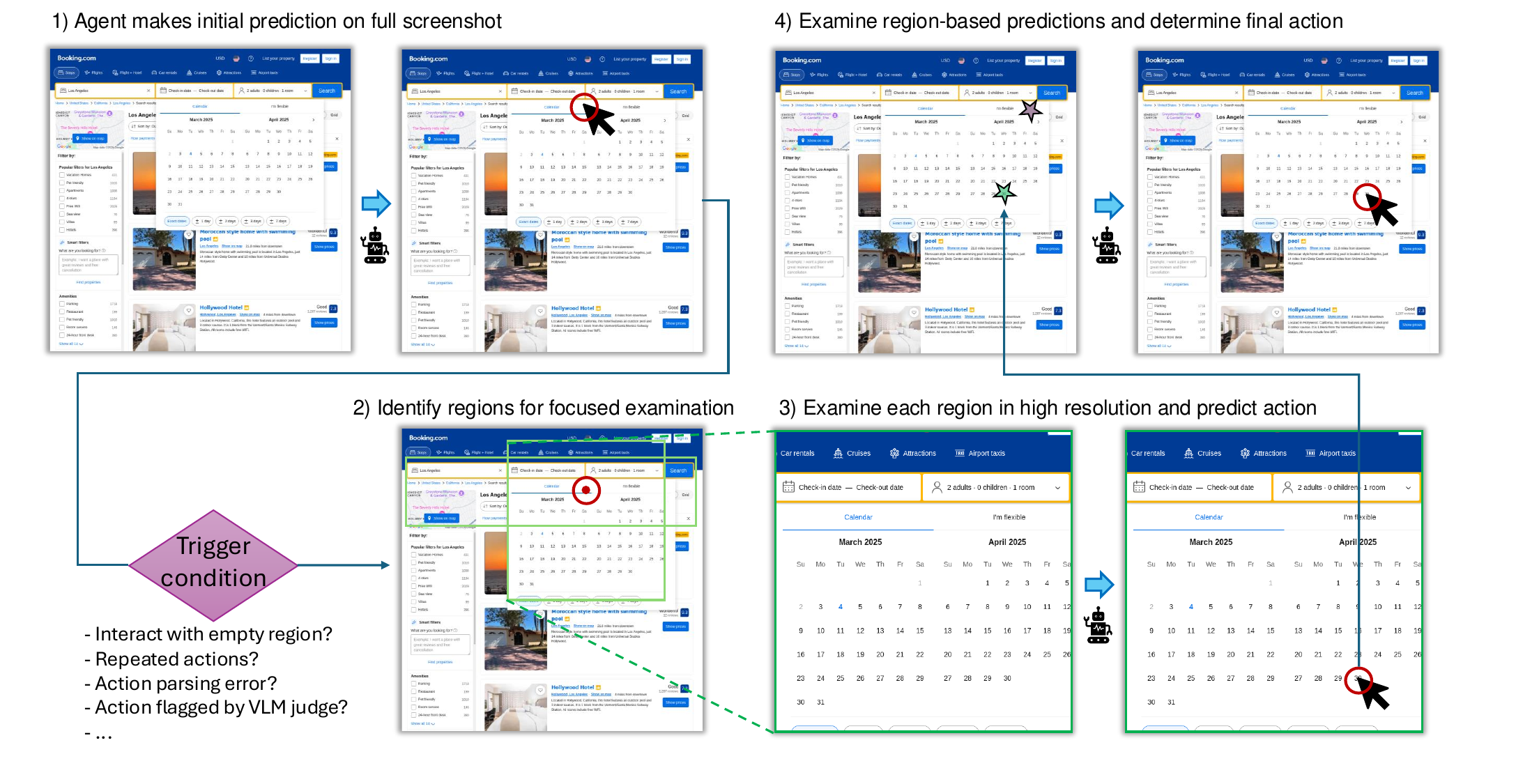}
    \vspace{-0.3in}
    \caption{\textbf{Overview of integrating RegionFocus into GUI agent pipelines.} The standard inference process (blue arrow) takes input information and continuously predicts subsequent actions. When the GUI agent encounters errors, RegionFocus (green arrow) activates, proposing focal points to extract targeted sub-regions. Actions are then predicted individually for these sub-regions and aggregated into a single refined action for standard inference. }
    \label{fig:overview}
\end{figure*}
\section{Related Work}
\label{sec:related-work}

\subsection{GUI Agents}
Recent advancements in Large Language Models (LLMs) and Vision Language Models (VLMs) have significantly enhanced GUI automation, enabling agents to effectively interact with diverse graphical environments through textual and visual modalities~\cite{yao2022webshop, deng2023mind2web, yan2023gpt, kim2023language, wang2024mobile, gou2024navigating, liu2024visualagentbench, wang2024mobileagent, pahuja2025explorerscalingexplorationdrivenweb, trabucco2025internetscaletrainingagents, murty2025nnetnavunsupervisedlearningbrowser}. Prior studies generally adopt two distinct approaches: (1) text-based reasoning, which leverages structured representations such as HTML or accessibility trees~\cite{koh2024visualwebarena, cao2025spider2, zhou2024webarena}, extracting structured interface information~\cite{lu2024omniparserpurevisionbased, yu2025omniparser} and supplementary textual details for input into LLMs/VLMs~\cite{zheng2024gpt}; and (2) vision-based inference, relying on VLMs to directly interpret GUI elements~\cite{gou2024navigating, qin2025ui}. While text-based techniques efficiently handle structured information, they often struggle with visually complex or ambiguous interfaces~\cite{xie2025osworld, he2024webvoyager, li2024screenspotpro}, resulting in inaccuracies and reduced reliability. Similar observations have been reported in \cite{yang2024agentoccamsimplestrongbaseline}.
 Conversely, visual grounding approaches may inadvertently interact with irrelevant or empty regions due to overly broad visual attention. In contrast to prior methods using entire interface screens as input~\cite{shaw2023pixels, cheng2024seeclick, hong2024cogagent}, our approach explicitly separates planning from visual grounding via a novel visual test-time scaling framework. This framework selectively targets salient GUI regions through precise bounding-box proposals and integrates an innovative ``image-as-map" strategy, maintaining contextual coherence throughout interactions.

\subsection{Test-Time Scaling in AI Agents}

Test-time scaling involves dynamically adjusting computational resources during inference to enhance model performance~\cite{wang2022self, yao2023react, yao2023tree, tran2025learning}. This approach allows AI agents to allocate additional processing power to challenging tasks, thereby improving decision-making and accuracy. Inspired by advancements in test-time scaling for LLMs, several studies have extended similar principles to GUI agents. For example, during the inference, \cite{zhang2023you} leverages intermediate action histories, \cite{nakano2021webgpt} collects external information during inference, and \cite{yu2024exact} incorporates reflection mechanisms into AI agents. Despite their success in improving performance, these methods do not utilize the unique advantages of visual information. In this paper, we propose a preliminary approach toward visual test-time scaling, dynamically adjusting the image focus region and employing an ``image-as-map" technique to encode historical information for more effective GUI agent inference.

\subsection{Visual Image Attention}

Visual image attention mechanisms have a rich history of enabling AI models to selectively focus on pertinent regions within visual inputs~\cite{xu2015show, linsley2018learning, farinhas2021multimodalcontinuousvisualattention, guo2022attention}. Such mechanisms are especially critical for GUI-based agents, where accurately identifying and interacting with interface elements amidst visually cluttered environments is essential. Our proposed approach introduces a region-focused mechanism utilizing predefined bounding boxes that progressively refine their attention through historical recording, incrementally concentrating focus more precisely on target elements. This iterative refinement shares conceptual similarities with prior recurrent models~\cite{mnih2014recurrent, cai2018cascade}; however, our method leverages VLMs to achieve refinement~\cite{nasiriany2024pivot}, uniquely integrating this process directly into GUI agents' test-time scaling. Accurately generating these attention regions directly via VLMs remains an open direction for future research.

\section{Method}
\label{sec:method}

To address the limitations of current GUI agent frameworks, we propose a visual test-time scaling approach that enhances the robustness and accuracy of VLM agents interacting with complex graphical user interfaces. Unlike traditional methods, which uniformly treat all interface elements, our method dynamically adjusts the model's focus by selectively emphasizing visually salient regions whenever potential errors are detected. This approach significantly reduces misclicks and navigation mistakes.

Crucially, our framework operates entirely at inference time, enabling straightforward integration into existing VLM agents without requiring retraining or architectural modifications. In the following sections, we first describe the integration process for our pipeline with existing GUI agents, followed by a detailed explanation of the design principles and implementation of each component.

\subsection{Overview}
\label{sec:method:overview}
Figure~\ref{fig:overview} illustrates our proposed pipeline. On top of the standard interaction pipeline, our approach introduces an error-triggered refinement mechanism. Specifically, when the agent encounters a \emph{trigger condition}—such as clicking on non-interactive elements (e.g., selecting empty space instead of the intended date option) or repeating unsuccessful actions (e.g., repeatedly typing ``Los Angeles" without successfully clicking the correct button)—the \emph{RegionFocus} component is activated.

During this refinement stage, the agent initially predicts a focal point near the intended target element. Based on this approximate location, it generates a bounding box likely to encapsulate the target element (Section~\ref{sec:method:region-focus}). For each region defined by the bounding boxes, the agent independently predicts candidate actions. Finally, the agent aggregates a single action to be executed based on the predictions for each region. 

Additionally, we maintain an annotated history of previously examined focal points on the same UI image using an \emph{image-as-map} representation (Section~\ref{sec:method:image-as-map}). This visual history guides the agent in avoiding redundant searches and helps it progressively focus on the correct target. Interactive predictions involving specific coordinates are visually marked (e.g., using pink-star landmarks) to help the model verify the correctness of its selections.

\subsection{Visual Region Focus}
\label{sec:method:region-focus}

The core of our pipeline is a dynamic visual adaptation mechanism activated during inference. When initial action predictions result in errors—such as clicks on non-interactive or empty regions—the model dynamically adjusts its visual attention. Specifically, it refines attention by generating bounding-box proposals around visually salient regions, leading to more precise single-step actions.

\paragraph{Trigger Condition} We define two primary types of triggering conditions. The first relies on environmental feedback obtained from direct interaction with dynamic GUI environments, such as interactive webpages. Errors, like clicks on non-interactive elements, can be easily detected based on environment feedback (e.g., webpage change). The second type of trigger occurs in cases where environmental interaction isn't possible and we are dealing with static screenshots (e.g., ScreenSpot-Pro~\cite{li2025screenspot} scenario). Here, the VLM evaluates predicted actions, shifting its role from merely making predictions to explicitly evaluating the correctness of actions.
Although the VLM's judgments aren't always perfectly accurate, they significantly help identify and mitigate errors shown in our experiments.

\paragraph{Bounding-box Proposal} Empirically, VLM agents reliably produce focal points near target elements but struggle to directly predict accurate bounding boxes~\cite{feng2025vision}. Instead, we derive bounding boxes from these focal points rather than predicting them directly. Although advanced segmentation models (e.g., SAM \citep{kirillov2023segment}) could provide more precise bounding boxes, we currently focus exclusively on leveraging the GUI agent itself. This ensures that future enhancements in GUI agents directly benefit our approach.

We employ a heuristic approach, defining bounding boxes with fixed dimensions and aspect ratios (e.g., $0.5$ width by $0.5$ height) centered around the focal point. If a bounding box exceeds the image boundary, it is adjusted to remain fully within the screenshot. This simple yet effective strategy strikes a good balance between accuracy and computational efficiency. 

\paragraph{Action Candidate Prediction}
Given the regions identified by the bounding boxes extracted from the previous stage, the agent predicts an action for each region. If the agent can interact with the environment (e.g., in the case of a webpage), a zoomed-in, high resolution view of the region is provided to the agent. In this case, at least one side of the region can be made to match the original full image resolution via zooming in. If environment interaction is not possible, we simply crop the region from the initial image and upsample it for prediction.

\paragraph{Action Aggregation}
After the GUI agent independently analyzes each bounding-box proposal and generates candidate actions, we select a single action to serve as the next step in the inference pipeline based on these candidates.
For coordinate-based actions (e.g., ``Click (x, y)" or ``Scroll (x, y) down"), we visually mark the candidate action coordinates on the snapshot, as shown in Figure~\ref{fig:teaser} (2).\footnote{Landmark annotations are only used for actions that involve interacting with a specific point and element in the current view.} This process significantly reduces the model’s workload by simplifying how textual coordinates are mapped and interpreted on the image. Empirically, we observe that incorporating these visual markers leads the model to select action candidates more accurately.

\subsection{Image-as-Map}
\label{sec:method:image-as-map}

\begin{figure}[!t]
    \centering
    \includegraphics[width=1\linewidth]{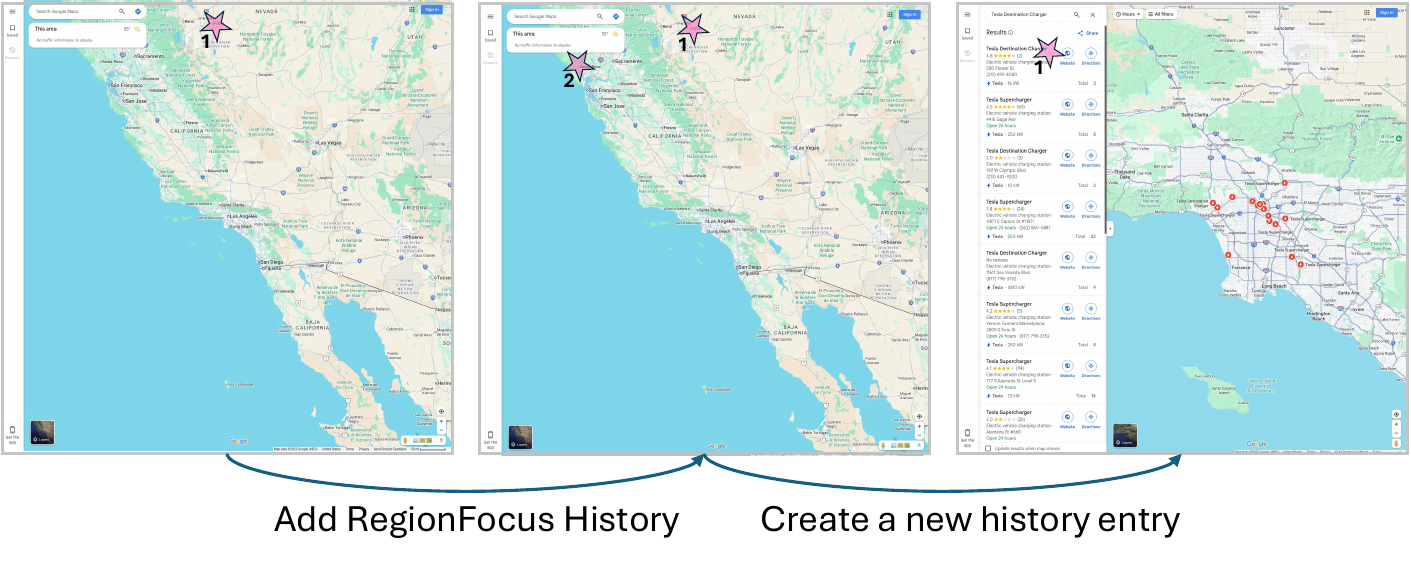}
    \vspace{-0.2in}
    \caption{\textbf{Image-as-Map records temporal information.} We place numbered pink stars in the image as visual landmarks to indicate previously focused points for the GUI agent. Each time a RegionFocus attempt fails (i.e., no action takes effect), we add a new pink star at the attempted location. Once an action successfully takes effect, we refresh the history and remove any existing landmarks.}
    \label{fig:method_RF}
\end{figure}

\begin{table*}[!ht]
\setlength{\tabcolsep}{3.5pt} %
\centering
\resizebox{\linewidth}{!}{
\begin{tabular}{l|ccc|ccc|ccc|ccc|ccc|ccc|ccc}
\toprule
\textbf{Agent Model} & \multicolumn{3}{c|}{\textbf{Development}} & \multicolumn{3}{c|}{\textbf{Creative}} & \multicolumn{3}{c|}{\textbf{CAD}} & \multicolumn{3}{c|}{\textbf{Scientific}} & \multicolumn{3}{c|}{\textbf{Office}} & \multicolumn{3}{c|}{\textbf{OS}} & \multicolumn{3}{c}{\textbf{Avg}} \\
\cmidrule(lr){2-4} \cmidrule(lr){5-7} \cmidrule(lr){8-10} \cmidrule(lr){11-13} \cmidrule(lr){14-16} \cmidrule(lr){17-19} \cmidrule(lr){20-22}
 & \textbf{text} & \textbf{icon} & \textbf{avg} & \textbf{text} & \textbf{icon} & \textbf{avg} & \textbf{text} & \textbf{icon} & \textbf{avg} & \textbf{text} & \textbf{icon} & \textbf{avg} & \textbf{text} & \textbf{icon} & \textbf{avg} & \textbf{text} & \textbf{icon} & \textbf{avg} & \textbf{text} & \textbf{icon} & \textbf{avg} \\
\midrule
QwenVL-7B & 0.0 & 0.0 & 0.0 & 0.0 & 0.0 & 0.0 & 0.0 & 0.0 & 0.0 & 0.7 & 0.0 & 0.4 & 0.0 & 0.0 & 0.0 & 0.0 & 0.0 & 0.0 & 0.1 & 0.0 & 0.1 \\
GPT-4o & 1.3 & 0.0 & 0.7 & 1.0 & 0.0 & 0.6 & 2.0 & 0.0 & 1.5 & 2.1 & 0.0 & 1.2 & 1.1 & 0.0 & 0.6 & 0.0 & 0.0 & 0.0 & 1.3 & 0.0 & 0.8 \\
SeeClick & 0.6 & 0.0 & 0.3 & 1.0 & 0.0 & 0.6 & 2.5 & 0.0 & 1.9 & 3.5 & 0.0 & 2.0 & 1.1 & 0.0 & 0.5 & 2.8 & 0.0 & 1.5 & 1.8 & 0.0 & 1.1 \\
Qwen2-VL-7B & 2.6 & 0.0 & 1.3 & 1.5 & 0.0 & 0.9 & 0.5 & 0.0 & 0.4 & 6.3 & 0.0 & 3.5 & 3.4 & 1.9 & 3.0 & 0.9 & 0.0 & 0.5 & 2.5 & 0.2 & 1.6 \\
OS-Atlas-4B & 7.1 & 0.0 & 3.7 & 3.0 & 1.4 & 2.3 & 2.0 & 0.0 & 1.5 & 9.0 & 5.5 & 7.5 & 5.1 & 3.8 & 4.4 & 5.6 & 0.0 & 3.1 & 5.0 & 1.7 & 3.7 \\
ShowUI-2B & 16.9 & 1.4 & 9.4 & 9.1 & 0.0 & 5.3 & 2.5 & 0.0 & 1.9 & 13.2 & 7.3 & 10.6 & 15.3 & 7.5 & 13.5 & 10.3 & 2.2 & 6.6 & 10.8 & 2.6 & 7.7 \\
CogAgent-18B & 14.9 & 0.7 & 8.0 & 9.6 & 0.0 & 5.6 & 7.1 & 3.1 & 6.1 & 22.2 & 1.8 & 13.4 & 13.0 & 0.0 & 6.5 & 5.6 & 0.0 & 3.1 & 12.0 & 0.8 & 7.7 \\
Aria-UI & 16.2 & 0.0 & 8.4 & 23.7 & 2.1 & 14.7 & 7.6 & 1.6 & 6.1 & 27.1 & 6.4 & 18.1 & 20.3 & 1.9 & 16.1 & 4.7 & 0.0 & 2.6 & 17.1 & 2.0 & 11.3 \\
UGround-7B & 26.6 & 2.1 & 14.7 & 27.3 & 2.8 & 17.0 & 14.2 & 1.6 & 11.1 & 31.9 & 2.7 & 19.3 & 31.6 & 11.3 & 27.9 & 17.8 & 0.0 & 9.7 & 25.0 & 2.8 & 16.5 \\
Claude Comp.Use & 22.0 & 3.9 & 12.6 & 25.9 & 3.4 & 16.8 & 14.5 & 3.7 & 11.9 & 33.9 & 15.8 & 25.8 & 30.1 & 16.3 & 26.2 & 11.0 & 4.5 & 8.1 & 23.4 & 7.1 & 17.1 \\
OS-Atlas-7B & 33.1 & 1.4 & 17.7 & 28.8 & 2.8 & 17.9 & 12.2 & 4.7 & 10.3 & 37.5 & 7.3 & 24.4 & 33.9 & 5.7 & 27.4 & 27.1 & 4.5 & 16.8 & 28.1 & 4.0 & 18.9 \\
UGround-V1-7B & - & - & 35.5 & - & - & 27.8 & - & - & 13.5 & - & - & 38.8 & - & - & 48.8 & - & - & 26.1 & - & - & 31.1 \\
\midrule
UI-TARS-7B & 58.4 & 12.4 & 36.1 & 50.0 & 9.1 & 32.8 & 20.8 & \textbf{9.4} & 18.0 & 63.9 & \textbf{31.8} & 50.0 & 63.3 & 20.8 & 53.5 & 30.8 & 16.9 & 24.5 & 47.8 & 16.2 & 35.7\\
+ \emph{RegionFocus} & \textbf{59.7} & \textbf{15.9} & \textbf{38.5} & \textbf{59.6} & \textbf{11.9} & \textbf{39.6} & \textbf{30.5} & 7.8 & \textbf{24.9} & \textbf{67.4} & 30.0 & \textbf{51.2} & \textbf{69.5} & \textbf{30.2} & \textbf{60.4} & \textbf{45.8} & \textbf{21.3} & \textbf{34.7} & \textbf{55.2} & \textbf{18.7} & \textbf{41.2}\\
UI-TARS-72B & 63.0 & 17.3 & 40.8 & 57.1 & 15.4 & 39.6 & 18.8 & 12.5 & 17.2 & 64.6 & 20.9 & 45.7 & 63.3 & 26.4 & 54.8 & 42.1 & 15.7 & 30.1 & 50.9 & 17.5 & 38.1 \\
+ \emph{RegionFocus} & \textbf{72.1} & \textbf{26.9} & \textbf{50.2} & \textbf{68.7} & \textbf{22.4} & \textbf{49.3} & \textbf{35.5} & \textbf{25.0} & \textbf{33.0} & \textbf{77.1} & \textbf{30.9} & \textbf{57.1} & \textbf{72.9} & \textbf{45.3} & \textbf{66.5} & \textbf{63.6} & \textbf{27.0} & \textbf{46.9} & \textbf{64.0} & \textbf{28.0} & \textbf{50.2}\\
\midrule
Qwen2.5-VL-7B & 45.5 & 1.4 & 24.1 & 32.8 & \textbf{6.3} & 21.7 & 22.3 & \textbf{6.2} & 18.4 & 50.7 & 7.3 & 31.9 & 52.5 & 15.1 & 43.9 & 36.4 & 10.1 & 24.5 & 39.3 & 6.6 & 26.8\\
+ \emph{RegionFocus} & \textbf{53.2} & \textbf{3.4} & \textbf{29.1} & \textbf{42.9} & 4.9 & \textbf{27.0} & \textbf{28.4} & 3.1 & \textbf{22.2} & \textbf{56.9} & \textbf{10.9} & \textbf{37.0} & \textbf{59.9} & \textbf{24.5} & \textbf{51.7} & \textbf{41.1} & \textbf{15.7} & \textbf{29.6} & \textbf{46.6} & \textbf{8.8} & \textbf{32.1}\\
Qwen2.5-VL-72B & 66.2 & 13.8 & 40.8 & 64.6 & 15.4 & 44.0 & 47.7 & 12.5 & 39.1 & 78.5 & 29.1 & 57.1 & 74.6 & 37.7 & 66.1 & 60.7 & 22.5 & 43.4 & 64.9 & 20.2 & 47.8\\
+ \emph{RegionFocus} & \textbf{75.3} & \textbf{25.5} & \textbf{51.2} & \textbf{76.3} & \textbf{30.8} & \textbf{57.2} & \textbf{71.6} & \textbf{28.1} & \textbf{60.9} & \textbf{87.5} & \textbf{39.1} & \textbf{66.5} & \textbf{87.0} & \textbf{60.4} & \textbf{80.9} & \textbf{74.8} & \textbf{36.0} & \textbf{57.1} & \textbf{78.6} & \textbf{34.1} & \textbf{61.6}\\
\bottomrule
\end{tabular}}
\vspace{-0.1in}
\caption{\textbf{Comparison of various models on ScreenSpot-Pro.}~\cite{li2025screenspot}. Our proposed RegionFocus helps the UI-TARS-72B~\cite{qin2025ui} model achieve a $31.8$\% improvement, while Qwen2.5-VL-72B~\cite{Qwen2.5-VL} sees a $28.9$\% gain, thereby achieving state-of-the-art performance. Additionally, integrating RegionFocus into UI-TARS-7B allows it to surpass the performance of the substantially larger UI-TARS-72B model.}
\label{tab:screenspot_pro_results}
\end{table*}

When RegionFocus is triggered multiple times for a given UI image, it is important for the agent to generate diverse focal points and avoid revisiting previously explored regions. 
Initially, we attempted to represent the region focus history using textual coordinates but found this approach ineffective, as the agent often revisited similar focal points despite explicit prompts to avoid them.

To address this, we propose an image-as-map representation, where we visually encode previously examined focal points as landmarks (e.g., pink stars) directly onto UI snapshots (Figure~\ref{fig:method_RF}) to record past action points. Unlike text-based histories, this visual representation more effectively conveys temporal information, allowing the agent to reason directly over the image and avoid revisiting the same areas. Note that the visual landmark annotation process used for action aggregation is also a form of the image-as-map strategy, capturing spatial information. Landmark-annotated snapshots are used only in the RegionFocus process, whereas the original inference pipeline (e.g., for the initial action prediction) receives unaltered UI images. We maintain these highlighted landmarks on the page until an action takes effect (i.e., causes a meaningful state change), at which point the history is refreshed. Our empirical results show that this image-as-map strategy consistently outperforms text-based methods, especially in more complex, multi-step GUI tasks. We also observe that it helps the agent distinguish between two GUI elements that are very close to each other, as shown in Figure~\ref{fig:ablation_Figures}~(2).

\section{Experiments}
\label{sec:exp}
\vspace{-0.1in}

In the experimental sections below, we integrate our proposed pipeline with UI-TARS~\cite{qin2025ui} and Qwen2.5-VL~\cite{Qwen2.5-VL}, two recently proposed GUI agent models that autonomously interacts with GUI screenshots, which have demonstrated exceptional performance in various GUI tasks. We evaluate our pipeline across both OS operation tasks and Web Interaction. For our main experiments, we adopt a fixed bounding box approach (Section~\ref{sec:method:region-focus}), using only the agent model itself. We also include ablation studies using SAM~\cite{kirillov2023segment}, which provides more accurate bounding boxes. Additional experimental details are provided in Appendix A.

\subsection{ScreenSpot-Pro} 
ScreenSpot-Pro \cite{li2025screenspot} is a recently introduced benchmark specifically designed to evaluate GUI grounding capabilities in complex, high-resolution professional desktop environments. These environments typically involve screenshots larger than $3k \times 2k$ (see Figure1(b) in their paper for specific configuration details). The benchmark’s emphasis on intricate, large-scale interfaces makes it an ideal platform for assessing our proposed pipeline, which incorporates a mechanism to zoom into local regions for more detailed analysis.

\begin{table*}[t]
\centering
\resizebox{\linewidth}{!}{
\begin{tabular}{@{}l|cccccccc@{}}
\toprule 
\textbf{Agent Model} & \textbf{Allrecipes} & \textbf{Amazon} & \textbf{Apple} & \textbf{ArXiv} & \textbf{GitHub} & \textbf{Booking} & \textbf{ESPN} & \textbf{Coursera} \\ \midrule
Claude \ \ & 45.9\%$_{\pm 3.4\%}$ & \ \  58.6\%$_{\pm 4.2\%}$  & \ \ 58.1\%$_{\pm 4.0\%}$  & \ \ 55.0\%$_{\pm 7.0\%}$ & \ \ 56.9\%$_{\pm 1.4\%}$ & \ \ 19.0\%$_{\pm 1.3\%}$  & \ \  46.2\%$_{\pm 1.3\%}$  & \ \  68.2\%$_{\pm 1.3\%}$   \\

GPT-4o & \ \ 56.3\%$_{\pm 1.3\%}$ & \ \  53.7\%$_{\pm 2.5\%}$  & \ \ 56.6\%$_{\pm 1.3\%}$  & \ \ 60.5\%$_{\pm 0.0\%}$ & \ \ 57.7\%$_{\pm 3.7\%}$ & \ \ 43.9\%$_{\pm 3.5\%}$  & \ \  44.0\%$_{\pm 2.7\%}$  & \ \  65.1\%$_{\pm 2.8\%}$   \\

WebVoyager & \ \ 51.1\%$_{\pm 2.2\%}$ & \ \  52.9\%$_{\pm 1.4\%}$  & \ \ 62.8\%$_{\pm 2.3\%}$  & \ \ 52.0\%$_{\pm 1.3\%}$ & \ \ 59.3\%$_{\pm 3.7\%}$ & \ \ 32.6\%$_{\pm 2.7\%}$  & \ \  47.0\%$_{\pm 1.3\%}$  & \ \  57.9\%$_{\pm 2.7\%}$   \\ 
\midrule
Qwen2.5-VL-7B & \ \ 47.2\%$_{\pm 4.4\%}$& \ \ 49.1\%$_{\pm 7.5\%}$& \ \ 47.3\%$_{\pm 2.5\%}$& \ \ 14.9\%$_{\pm 2.1\%}$& \ \ 23.9\%$_{\pm 4.8\%}$& \ \ 10.0\%$_{\pm 0.9\%}$& \ \ 39.2\%$_{\pm 7.2\%}$& \ \ 46.4\%$_{\pm 0.3\%}$\\
+ \emph{RegionFocus} & \ \ \textbf{49.2}\%$_{\pm 7.2\%}$ & \ \ \textbf{53.6}\%$_{\pm 4.4\%}$ & \ \ \textbf{67.1}\%$_{\pm 3.6\%}$ & \ \ \textbf{51.7}\%$_{\pm 1.3\%}$ & \ \ \textbf{35.0}\%$_{\pm 0.3\%}$ & \ \ \textbf{30.0}\%$_{\pm 0.\%}$ & \ \ \textbf{39.6}\%$_{\pm 6.4\%}$ & \ \ \textbf{71.1}\%$_{\pm 4.8\%}$\\
Qwen2.5-VL-72B & \ \ 28.6\%$_{\pm 1.5\%}$& \ \ 56.2\%$_{\pm 4.3\%}$& \ \ 53.0\%$_{\pm 0.9\%}$& \ \ 32.6\%$_{\pm 2.4\%}$& \ \ 61.5\%$_{\pm 4.3\%}$& \ \ 24.8\%$_{\pm 2.9\%}$& \ \ 49.8\%$_{\pm 3.8\%}$& \ \ 72.9\%$_{\pm 1.9\%}$\\
+ \emph{RegionFocus} & \ \ \textbf{43.4}\%$_{\pm 1.3\%}$ & \ \ \textbf{60.6}\%$_{\pm 1.5\%}$ & \ \ \textbf{69.7}\%$_{\pm 4.0\%}$ & \ \ \textbf{45.4}\%$_{\pm 2.0\%}$ & \ \ \textbf{67.3}\%$_{\pm 3.8\%}$ & \ \ \textbf{37.3}\%$_{\pm 2.9\%}$ & \ \ \textbf{59.8}\%$_{\pm 1.4\%}$ & \ \ \textbf{78.2}\%$_{\pm 1.4\%}$\\
\midrule
UI-TARS-7B & \ \ 17.8\%$_{\pm 1.3\%}$ & \ \ 30.9\%$_{\pm 1.4\%}$& \ \ 17.1\%$_{\pm 1.3\%}$& \ \ 20.9\%$_{\pm 2.3\%}$& \ \ 32.5\%$_{\pm 2.5\%}$& \ \ 7.6\%$_{\pm 1.3\%}$& \ \ 45.0\%$_{\pm 1.3\%}$& \ \ 70.7\%$_{\pm 1.2\%}$\\
+ \emph{RegionFocus} & \ \ \textbf{35.6}\%$_{\pm 2.2\%}$
& \ \ \textbf{39.0}\%$_{\pm 2.4\%}$
& \ \ \textbf{31.8}\%$_{\pm 2.7\%}$
& \ \ \textbf{37.2}\%$_{\pm 0.0\%}$
& \ \ \textbf{58.1}\%$_{\pm 1.5\%}$
& \ \ \textbf{15.2}\%$_{\pm 1.3\%}$
& \ \ \textbf{62.8}\%$_{\pm 0.0\%}$
& \ \ \textbf{74.2}\%$_{\pm 1.4\%}$\\
UI-TARS-72B & \ \ 16.2\%$_{\pm 2.6\%}$& \ \ 38.6\%$_{\pm 2.0\%}$& \ \ 45.9\%$_{\pm 1.2\%}$& \ \ \textbf{64.9}\%$_{\pm 4.1\%}$& \ \ 43.2\%$_{\pm 2.0\%}$& \ \ 40.2\%$_{\pm 2.3\%}$& \ \ 47.2\%$_{\pm 3.8\%}$& \ \ 58.9\%$_{\pm 1.6\%}$\\
+ \emph{RegionFocus} & \ \ \textbf{57.5}\%$_{\pm 1.1\%}$
& \ \ \textbf{55.8}\%$_{\pm 1.4\%}$
& \ \ \textbf{52.6}\%$_{\pm 1.3\%}$
& \ \ 50.3\%$_{\pm 3.6\%}$
& \ \ \textbf{58.9}\%$_{\pm 3.7\%}$
& \ \ \textbf{44.7}\%$_{\pm 2.6\%}$
& \ \ \textbf{69.7}\%$_{\pm 1.3\%}$
& \ \ \textbf{82.2}\%$_{\pm 1.5\%}$\\
\midrule \midrule
\multirow{2}{*}{\textbf{Agent Model}} & \textbf{Cambridge} & \textbf{BBC} & \textbf{Google} & \textbf{Google} & \textbf{Google} & \multirow{2}{*}{\textbf{Huggingface}} & \textbf{Wolfram} & \multirow{2}{*}{\textbf{Overall}}  \\ 
& \textbf{Dictionary} & \textbf{News} & \textbf{Flights} & \textbf{Map} & \textbf{Search} & & \textbf{Alpha} & \\ 
\midrule 
Claude & \ \ 71.3\%$_{\pm 3.6\%}$ & \ \  66.7\%$_{\pm 4.8\%}$  & \ \ 15.1\%$_{\pm 5.5\%}$  & \ \ 55.3\%$_{\pm 1.4\%}$ & \ \ 72.9\%$_{\pm 1.3\%}$ & \ \ 53.5\%$_{\pm 4.7\%}$  & \ \  51.5\%$_{\pm 5.4\%}$  & \ \  52.8\%$_{\pm 1.4\%}$  \\ 
GPT-4o & \ \ 82.2\%$_{\pm 1.3\%}$ & \ \  54.8\%$_{\pm 2.4\%}$  & \ \ 28.6\%$_{\pm 0.0\%}$  & \ \ 56.9\%$_{\pm 2.8\%}$ & \ \ 63.6\%$_{\pm 1.3\%}$ & \ \ 42.6\%$_{\pm 3.6\%}$  & \ \  65.2\%$_{\pm 2.2\%}$  & \ \  55.5\%$_{\pm 0.8\%}$  \\
WebVoyager & \ \ 71.3\%$_{\pm 1.3\%}$ & \ \  60.3\%$_{\pm 2.8\%}$  & \ \ 51.6\%$_{\pm 1.4\%}$  & \ \ 64.3\%$_{\pm 2.8\%}$ & \ \ 77.5\%$_{\pm 2.7\%}$ & \ \ 55.8\%$_{\pm 2.3\%}$  & \ \  60.9\%$_{\pm 2.2\%}$  & \ \  57.1\%$_{\pm 0.2\%}$  \\ 
\midrule
Qwen2.5-VL-7B  & \ \ \textbf{21.1}\%$_{\pm 4.8\%}$& \ \ 45.1\%$_{\pm 3.0\%}$& \ \ 10.0\%$_{\pm 0.9\%}$& \ \ \textbf{30.2}\%$_{\pm 1.5\%}$& \ \ 10.0\%$_{\pm 1.3\%}$& \ \ 41.4\%$_{\pm 0.3\%}$& \ \ \textbf{51.3}\%$_{\pm 3.8\%}$& \ \ 32.5\%$_{\pm 1.3\%}$\\
+ \emph{RegionFocus} & \ \ 17.3\%$_{\pm 0.3\%}$ & \ \ \textbf{52.9}\%$_{\pm 0.9\%}$ & \ \ \textbf{12.8}\%$_{\pm 4.8\%}$ & \ \ 17.1\%$_{\pm 1.2\%}$ & \ \ \textbf{18.3}\%$_{\pm 2.4\%}$ & \ \ \textbf{60.0}\%$_{\pm 2.8\%}$ & \ \ 40.8\%$_{\pm 1.2\%}$ & \ \ \textbf{41.1}\%$_{\pm 1.2\%}$\\
Qwen2.5-VL-72B & \ \ 63.7\%$_{\pm 1.3\%}$& \ \ 45.9\%$_{\pm 1.6\%}$& \ \ 17.7\%$_{\pm 1.5\%}$& \ \ 31.2\%$_{\pm 1.5\%}$& \ \ 11.5\%$_{\pm 0.0\%}$& \ \ 38.3\%$_{\pm 2.9\%}$& \ \ 48.9\%$_{\pm 2.5\%}$& \ \ 42.4\%$_{\pm 0.5\%}$\\
+ \emph{RegionFocus} & \ \ \textbf{68.9}\%$_{\pm 1.3\%}$ & \ \ \textbf{54.4}\%$_{\pm 2.9\%}$ & \ \ \textbf{34.6}\%$_{\pm 4.4\%}$ & \ \ \textbf{42.2}\%$_{\pm 1.5\%}$ & \ \ \textbf{20.3}\%$_{\pm 0.0\%}$ & \ \ \textbf{51.5}\%$_{\pm 4.9\%}$ & \ \ \textbf{56.0}\%$_{\pm 1.1\%}$ & \ \ \textbf{52.7}\%$_{\pm 1.1\%}$\\
\midrule
UI-TARS-7B  & \ \ \textbf{57.1}\%$_{\pm 1.3\%}$& \ \ 41.3\%$_{\pm 2.7\%}$& \ \ 10.3\%$_{\pm 1.4\%}$& \ \ 17.5\%$_{\pm 0.0\%}$& \ \ 49.6\%$_{\pm 1.3\%}$& \ \ 40.7\%$_{\pm 1.4\%}$& \ \ 38.4\%$_{\pm 1.3\%}$& \ \ 33.2\%$_{\pm 0.5\%}$\\
+ \emph{RegionFocus} & \ \ 55.8\%$_{\pm 0.0\%}$
& \ \ \textbf{52.4}\%$_{\pm 2.4\%}$
& \ \ \textbf{28.6}\%$_{\pm 4.8\%}$
& \ \ \textbf{29.3}\%$_{\pm 2.4\%}$
& \ \ \textbf{60.5}\%$_{\pm 2.3\%}$
& \ \ \textbf{45.7}\%$_{\pm 1.3\%}$
& \ \ \textbf{44.2}\%$_{\pm 1.3\%}$
& \ \ \textbf{44.7}\%$_{\pm 0.5\%}$\\
UI-TARS-72B & \ \ 72.9\%$_{\pm 1.4\%}$& \ \ 39.3\%$_{\pm 3.4\%}$& \ \ 33.9\%$_{\pm 2.1\%}$& \ \ 27.2\%$_{\pm 5.2\%}$& \ \ 60.6\%$_{\pm 1.2\%}$& \ \ 24.9\%$_{\pm 1.6\%}$& \ \ 48.3\%$_{\pm 2.3\%}$& \ \ 44.1\%$_{\pm 0.5\%}$\\
+ \emph{RegionFocus} & \ \ \textbf{75.9}\%$_{\pm 1.3\%}$
& \ \ \textbf{49.8}\%$_{\pm 3.6\%}$
& \ \ \textbf{49.9}\%$_{\pm 2.1\%}$
& \ \ \textbf{59.9}\%$_{\pm 1.4\%}$
& \ \ \textbf{65.8}\%$_{\pm 1.2\%}$
& \ \ \textbf{59.6}\%$_{\pm 1.3\%}$
& \ \ \textbf{60.2}\%$_{\pm 1.3\%}$
& \ \ \textbf{59.5}\%$_{\pm 0.1\%}$\\
\bottomrule 
\end{tabular}
}
\vspace{-0.1in}
\caption{\textbf{Comparison of various models on WebVoyager}~\cite{he2024webvoyager}. For each automatic evaluation, we run GPT evaluator three times to calculate the performance mean and standard deviation. We evaluated our method on UI-TARS~\cite{qin2025ui} and Qwen2.5-VL~\cite{Qwen2.5-VL}, consistently observing performance improvements.} 
\label{tab:webvoyager_results}
\end{table*}

Specifically, ScreenSpot-Pro comprises expert-annotated tasks across 23 applications spanning five domains and three operating systems, thereby providing an extensive assessment of model performance. Tasks are categorized by functional domains, including Development, Creative, CAD, Scientific, Office, and Operating Systems, and are further divided into text-based and icon/widget-based grounding challenges. This structure facilitates a nuanced evaluation of grounding capabilities, particularly in tasks that require precise localization and interaction with small or visually similar GUI elements. The benchmark enforces stringent metrics, measuring grounding accuracy based on whether the model-predicted location falls within the bounding box of the target element.

Because ScreenSpot-Pro only provides static screenshots—without an OS environment for interaction—we employ a VLM-based judge to trigger RegionFocus. Specifically, when the agent predicts a point, we will highlight that point in the input screenshot with pink-star landmarks and ask the model itself to assess the correctness of that point. If deemed incorrect, we initiate the RegionFocus process.

\paragraph{Results}
In Table~\ref{tab:screenspot_pro_results}, we summarize the reported grounding accuracy of various methods evaluated on ScreenSpot-Pro. For fair comparisons, we employ the 
\href{https://github.com/likaixin2000/ScreenSpot-Pro-GUI-Grounding}{official test code} 
released by ScreenSpot-Pro for evaluation. We report the original numbers from the UI-TARS paper in Table~\ref{tab:screenspot_pro_results}.

From Table~\ref{tab:screenspot_pro_results}, we can see that RegionFocus consistently improves performance across all categories for both text and icon grounding when compared to the base model. UI-TARS-72B + RegionFocus achieves a $31.7$\% improvement over the base UI-TARS-72B model. Moreover, the UI-TARS-7B + RegionFocus variant outperforms the UI-TARS-72B model overall, demonstrating the effectiveness of our approach. Furthermore, RegionFocus further helps QWen2.5-VL-72B achieve the state-of-the-art performance, $61.6$\%.

\begin{figure}[!t]
    \centering
    \includegraphics[width=1\linewidth]{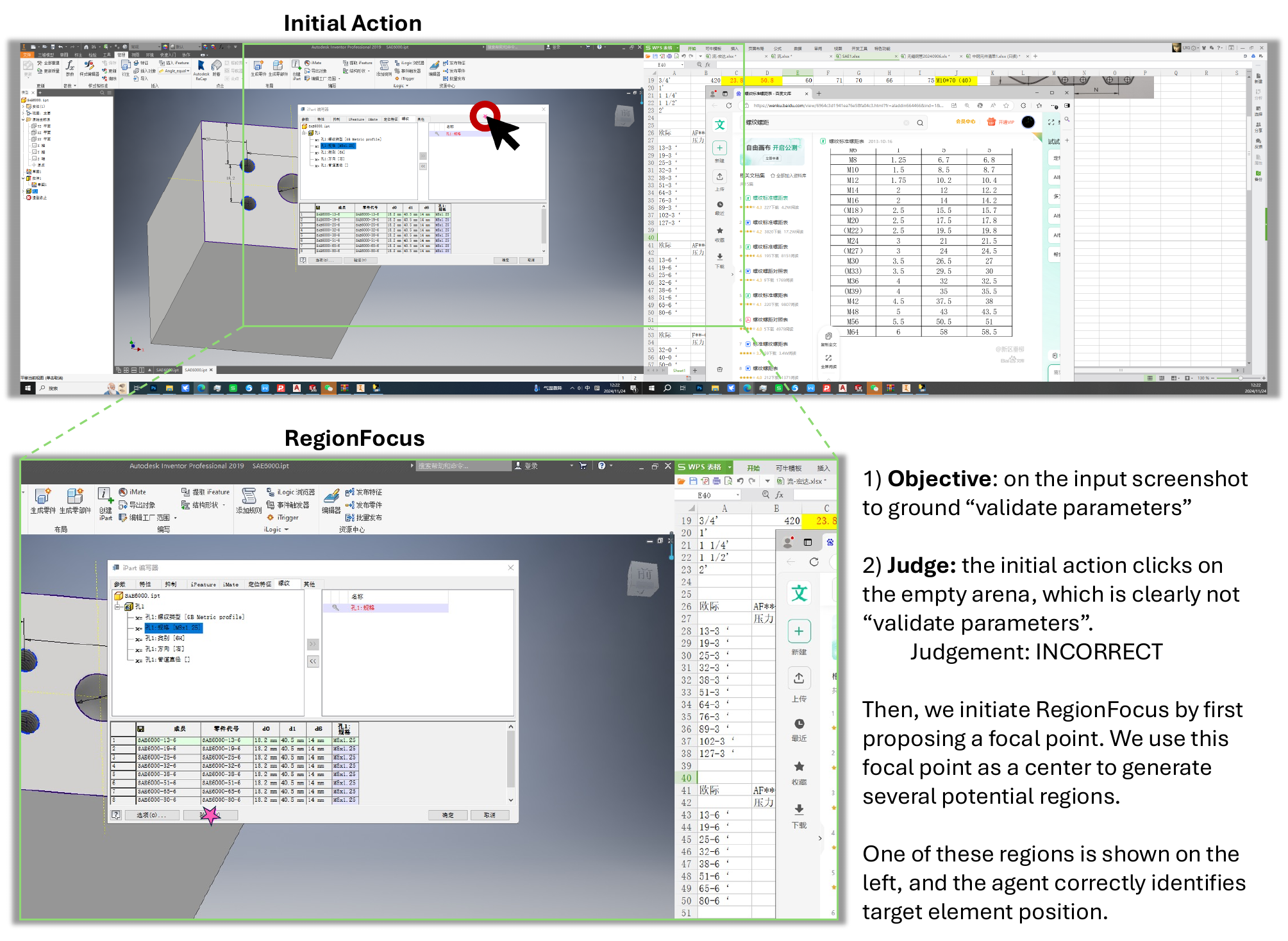}
    \caption{\textbf{Qualitative results - Screenspot-Pro.} In one example from our evaluation, the agent successfully rejects the initial action via self-VLM-judge and proposes a correct grounding point based on the zoomed-in view. More qualitative results are listed in Appendix B.}
    \label{fig:qual-screenspot}
\end{figure}

\begin{figure*}[t]
    \centering
    \includegraphics[width=1\linewidth,trim={1.5em 48em 2.0em 0},clip]{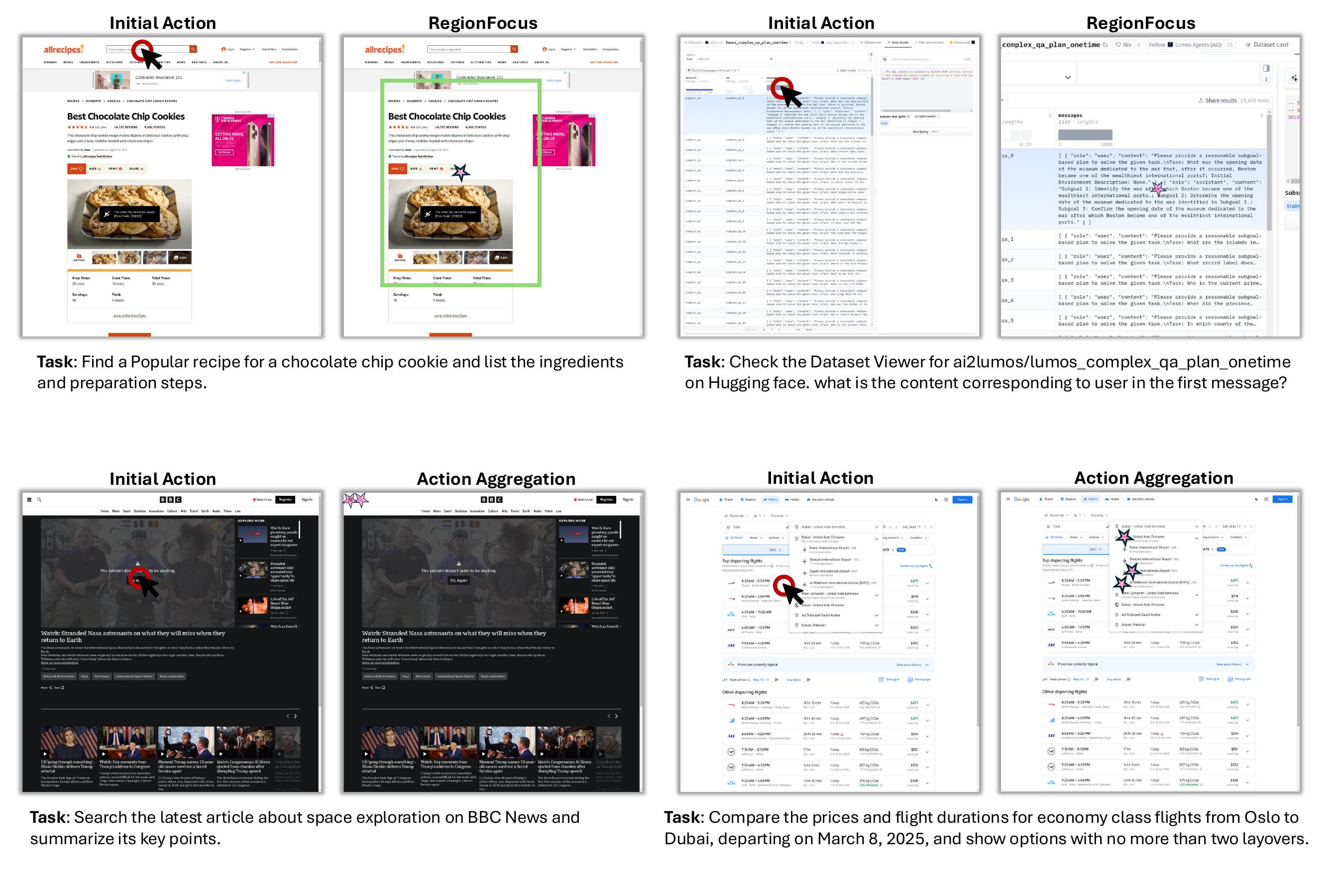}
    \vspace{-0.2in}
    \caption{\textbf{Qualitative Results - RegionFocus.} In these two examples, we illustrate how RegionFocus reduces background noise by emphasizing salient regions of an image. The mouse pointer indicates the agent’s initial action prediction, which is suboptimal in both cases. \textbf{Left pair of images}: The green window in the second image marks the zoomed-in region. By focusing on this region, we naturally cut out the distracting portion of the first image. \textbf{Right pair of images}: The second image is zoomed in, significantly reducing distracting details. This allows the agent to focus on the relevant information—even though the distracting region from the first image is still visible.}
    \label{fig:qual-webvoyager1}
\end{figure*}

\begin{figure*}[t]
    \centering
    \includegraphics[width=1\linewidth,trim={1.5em 4em 2.0em 45em},clip]{images/qual_webvoyager_v2.pdf}
    \vspace{-0.2in}
    \caption{\textbf{Qualitative results - image-as-map.} These examples demonstrate how action aggregation, enhanced by the proposed image-as-map, helps distinguish subtle coordinate differences between target elements. The mouse pointer indicates the agent’s initial predictions, which were incorrect in both cases. Each star-like landmark is generated during the RegionFocus process before action aggregation. \textbf{Left pair of images}: The two landmarks at the top left correspond to the home and search buttons. \textbf{Right pair of images}: the landmarks correspond to different options in a dropdown menu.
    }
    \label{fig:qual-webvoyager2}
\end{figure*}

Figure~\ref{fig:qual-screenspot} illustrates how our pipeline works in ScreenSpot-Pro: we first ask the agent to judge the initial action prediction. If it is incorrect, we initiate RegionFocus by zooming into the regions predicted by the agent, predicting actions within each region, and finally aggregating all actions into a single outcome. Because we used the agent model itself to judge whether a prediction is correct or incorrect, our results show that although the VLM may generate incorrect coordinates initially, it can still reliably judge whether those click points are correct with the help of image-as-map. Then, such an incorrect prediction can be corrected by the RegionFocus process. This generation-verification gap has also been noted in recent literature~\cite{cook2023complexity, swamy2025all}.

\subsection{WebVoyager}
\label{sec:exp:webvoyager}
WebVoyager \cite{he2024webvoyager} is a benchmark designed to evaluate autonomous web agents' capabilities in performing complex, open-ended tasks through multimodal interactions with real-world websites. Distinct from previous web agent benchmarks, WebVoyager comprises 643 semi-automatically generated tasks across 15 popular, real-world websites such as Amazon, Apple, ArXiv, and Google Maps. This selection ensures a diverse range of interactions reflecting everyday web browsing scenarios. Tasks in WebVoyager require agents to process visual information from rendered screenshots and textual cues from web elements, enabling nuanced evaluation of multimodal reasoning and navigation skills. Furthermore, the benchmark introduces an automatic evaluation protocol utilizing GPT-4V, achieving 85.3\% agreement with human judgment, thereby offering a reliable assessment of agent performance. 

In this scenario, the agent actively interacts with the web environment. We employ a Playwright-controlled Chrome browser to navigate webpages, with the VLM agent determining the appropriate action based on each webpage screenshot. After task execution, we use the official evaluation setting, where a GPT-based judge reviews the last 15 screenshots along with an optional textual response to determine whether the task has been successfully accomplished.

\begin{figure}[!t]
    \centering
    \includegraphics[width=1\linewidth]{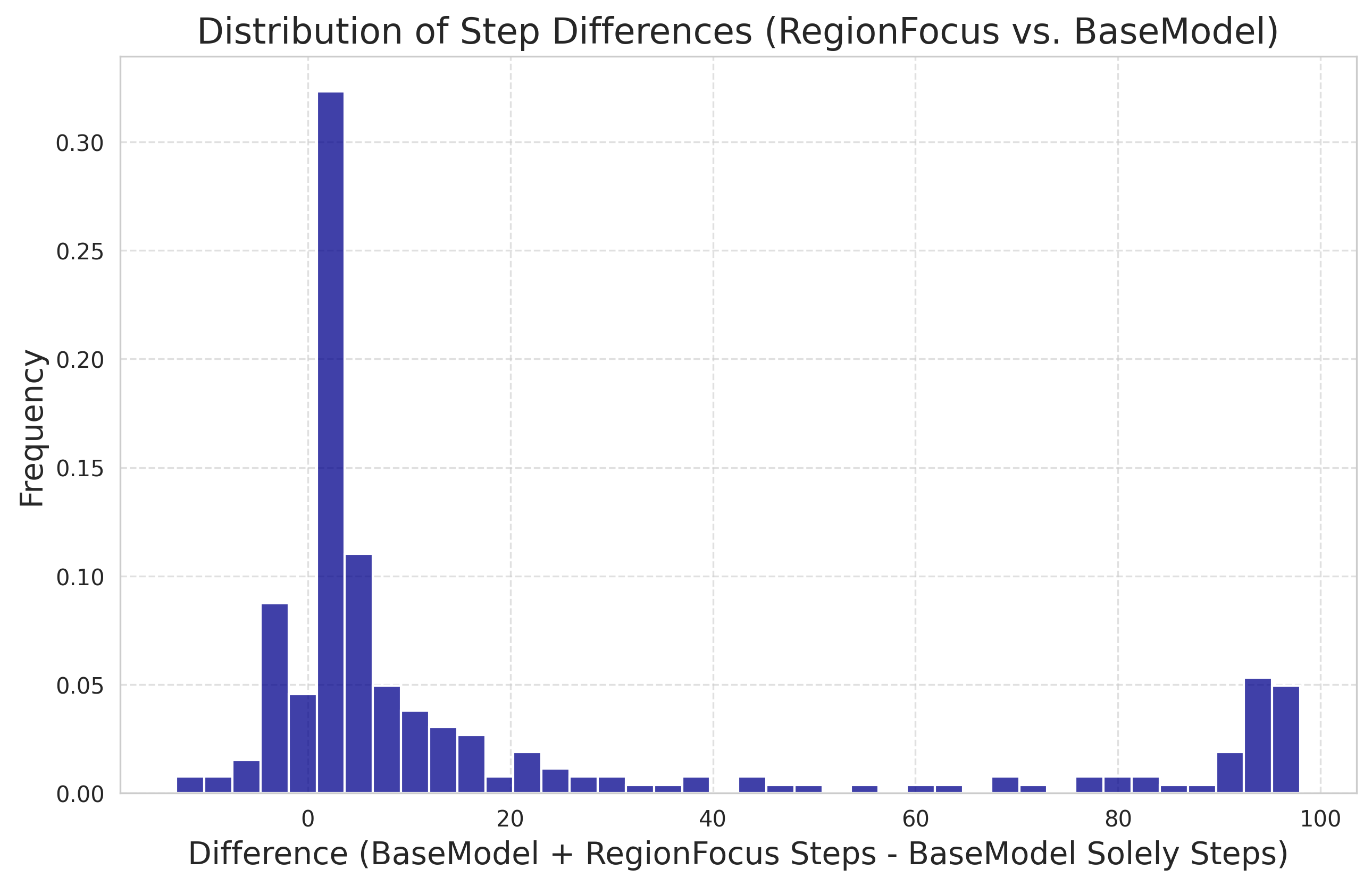}
    \vspace{-0.2in}
    \caption{\textbf{Histogram of step differences}: BaseModel + RegionFocus vs. BaseModel alone. BaseModel is UI-TARS-72B.}
    \label{fig:histogram}
\end{figure}

\paragraph{Results}
For comparative analysis, Table \ref{tab:webvoyager_results} presents the task success rates of various web agents evaluated on the WebVoyager benchmark. RegionFocus consistently improves performance across all types of websites—including ``Booking" and ``Search"—highlighting the effectiveness of integrating RegionFocus into the GUI agent for web browsing. It also brings consistent improvements over two open-source model, UI-TARS and Qwen2.5-VL. Please note that our model performance was impacted by online interaction constraints—such as bot blocking and intermittent VPN issues. By resolving these factors, we can further boost the model’s overall performance. 

We present several qualitative examples of WebVoyager’s performance in Figures~\ref{fig:qual-webvoyager1},~\ref{fig:qual-webvoyager2}. In Figure~\ref{fig:qual-webvoyager1} left, the agent initially fails by clicking the `ingredients' button, which appears in the search bar despite being on the correct page. By highlighting the relevant region with a green bounding box, RegionFocus naturally filters out background noise and draws attention to the primary content. In Figure~\ref{fig:qual-webvoyager1} right, RegionFocus zooms in on the sub-region of interest, enlarging key content and making it easier for the agent to locate the target content. Figure~\ref{fig:qual-webvoyager2} left shows a case where the agent initially clicks an unrelated element. Our pipeline then corrects this mistake by proposing two closely positioned buttons. The image-as-map mechanism allows the agent to distinguish between these nearly identical elements, even when their coordinates differ only slightly.  Finally, Figure~\ref{fig:qual-webvoyager2} right illustrates a scenario where the agent mistakenly clicks on an empty area close to the desired element. Once again, RegionFocus highlights the correct button, helping the agent choose it accurately.

\paragraph{More Analysis} Figure~\ref{fig:histogram} shows the distribution of step differences between the combined BaseModel + RegionFocus approach and the BaseModel alone over 400 trajectories. Only the actual browser-interactive steps are counted, excluding RegionFocus overhead. Here the BaseModel is UI-TARS-72B. As shown, BaseModel + RegionFocus generally yields more steps on average ($19.74$\% steps), correlating with a overall $34.3$\% higher success rate. On average, RegionFocus is triggered $5.8$ times per Web Browsing trajectory. Furthermore, in $32.3$\% of cases RegionFocus is triggered only once, yet a single trigger yields an impressive $83.7$\% increase in the success of those trajectories.

\begin{figure}[!t]
    \centering
    \includegraphics[width=1.0\linewidth]{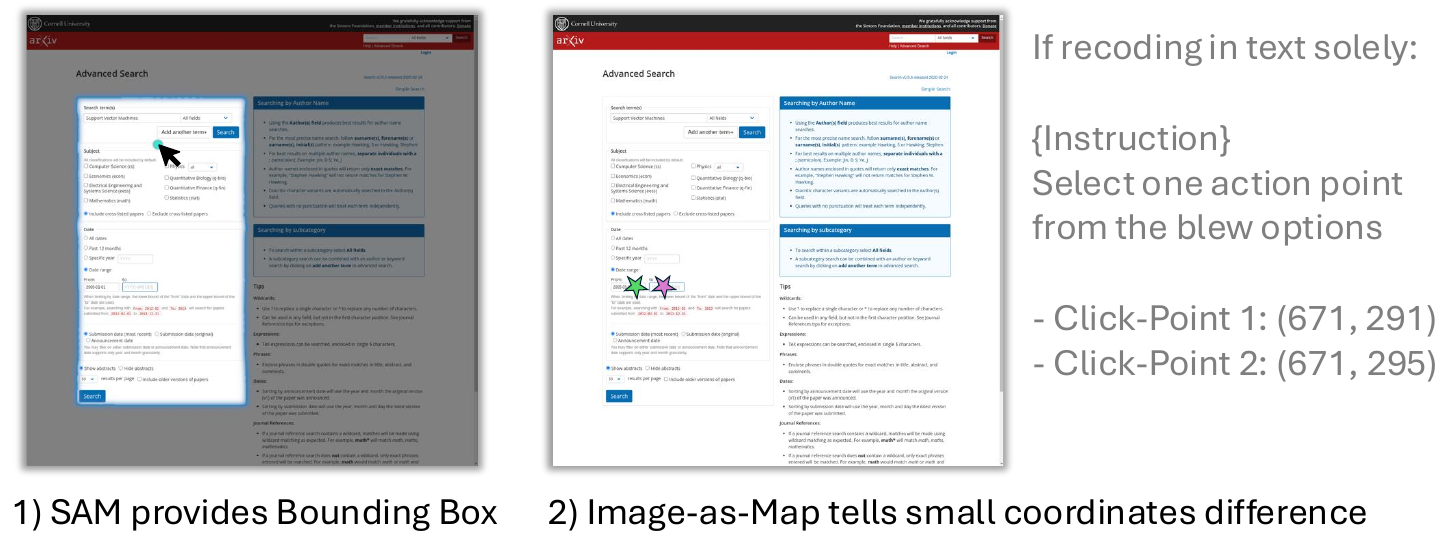}
    \vspace{-0.2in}
    \caption{\textbf{Ablation studies.} (1) RegionFocus can natively use the SAM to generate bounding boxes. (2) how image-as-map helps highlight subtle differences of different GUI elements.}
    \label{fig:ablation_Figures}
\end{figure}

\subsection{Ablation Studies}
\label{sec:exp:ablation}
\begin{wraptable}{r}{0.2\textwidth}
\small
\centering
\setlength{\tabcolsep}{1.6mm}{
\begin{tabular}{@{}l|c@{}}
\toprule 
\textbf{Agent Model} & Overall \\
\midrule
image-as-map &  43.2\\
Text-as-History & 37.2\\
\midrule
Fixed-BBox & 43.2 \\ 
Predict-Region & 28.1\\
SAM & 46.5\\
\bottomrule 
\end{tabular}
}
\caption{\textbf{Ablation study results.} We tested on a subset of WebVoyager, and the score is higher the better.} 
\label{tab:ablation_results}
\end{wraptable}
We conduct ablation studies on the entire pipeline, including our ``image-as-map" design choice and the use of a predefined bounding box based on the point predicted by the agent. We also demonstrate that by leveraging SAM~\cite{kirillov2023segment} and increasing the number of trajectory steps, performance can be further improved. For the sake of computation, we employed the UI-TARS-7B-DPO model for these ablation studies on a subset of the WebVoyager benchmark. The results are shown in Table~\ref{tab:ablation_results}, where ``image-as-map" and ``Fixed-BBox" refers to our same 7B model configuration with RegionFocus enabled and a maximum limit of 100 action steps.

\paragraph{Text-based RegionFocus History Representation}
By using image-as-map, we can directly provide visual location information to the agent, helping it distinguish even minor differences and thereby enhancing its perception. For instance, as shown in Figure~\ref{fig:ablation_Figures} (2), we color-code click points in the image to denote the image-as-map mechanism, while the corresponding coordinates are listed in the text box on the right. Notably, although the textual coordinate difference is within only five pixels, the resulting action can vary significantly. In this ablation study, we use a text representation for both history tracking and action aggregation, which leads to significant degradation compared to our image-as-map representation.

\paragraph{Proposing regions directly}
In our pipeline (Section~\ref{sec:method}), rather than having the agent model directly propose a bounding box, we first prompt it to identify a point of interest and then generate a predefined bounding box around that point. This design choice stems from the observation that agent models often struggle to accurately predict bounding boxes on their own. To validate this, we conducted an experiment in which the model was required to predict both the upper-left and bottom-right corner coordinates, which were then used to crop the UI image. As shown by the ``Predict-Region" results, this approach led to a marked decrease in performance.

\paragraph{SAM}
As discussed in Section~\ref{sec:method}, our pipeline can naturally leverage segmentation models that take point inputs as indicator, such as SAM~\cite{kirillov2023segment}. For example, in Figure~\ref{fig:ablation_Figures} (1), we provide a point generated by the model for RegionFocus, despite the fact that the point itself is referring to a non-interactive empty area. Nevertheless, SAM is able to produce a bounding box that includes the correct region, showcasing its suitability in such cases. The effectiveness of incorporating point-based segmentation models into RegionFocus is further validated by the results in Table~\ref{tab:ablation_results}.

\begin{figure}[!t]
    \centering
    \includegraphics[width=1\linewidth]{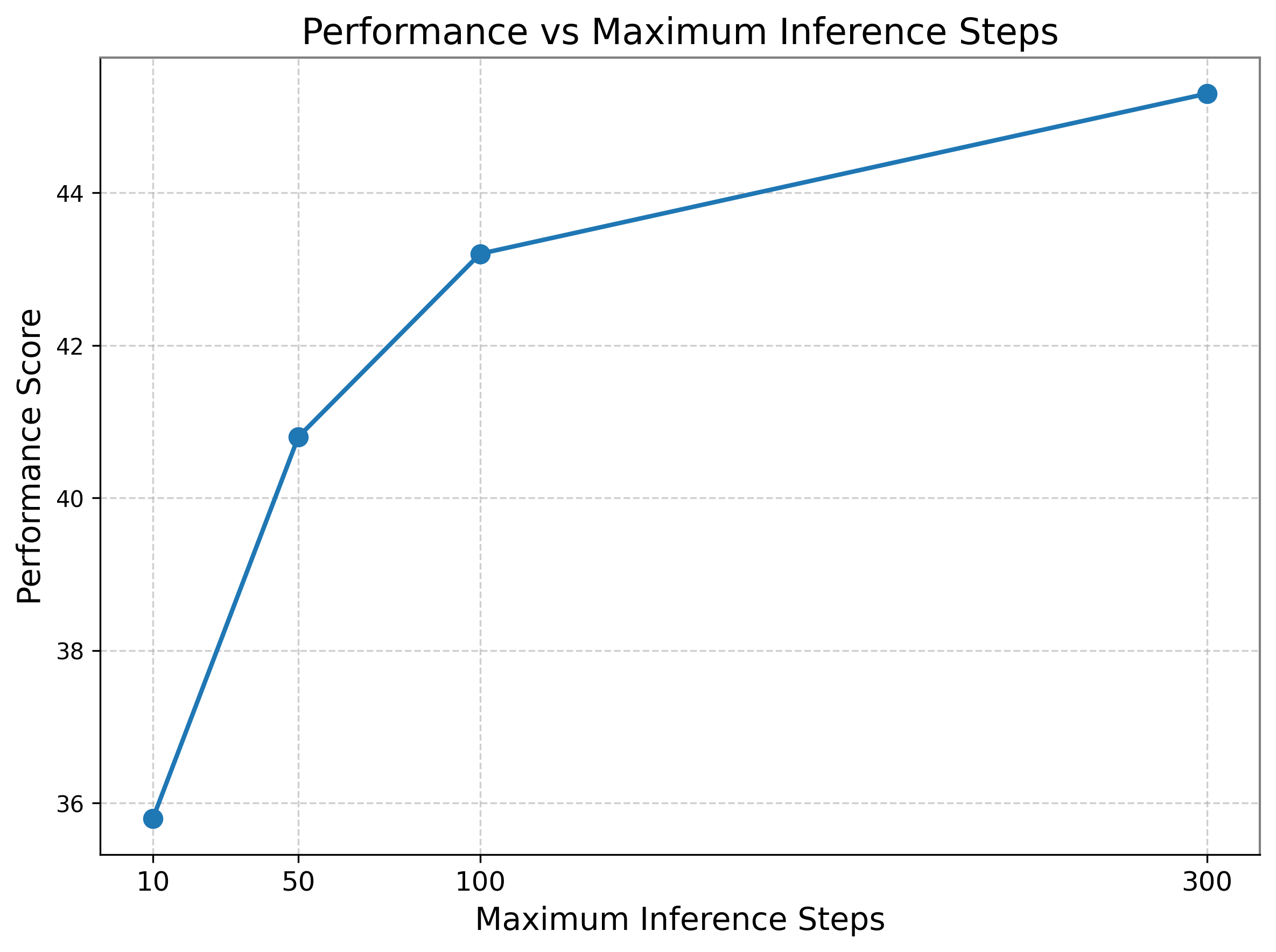}
    \vspace{-0.2in}
    \caption{\textbf{Ablation study -- inference steps}: higher limits yield further improvements, though the benefits gradually decay.}
    \label{fig:ablation_steps}
\end{figure}

\paragraph{Test-time Thinking Budget}
Our quantitative analysis in Figure~\ref{fig:histogram} shows that incorporating RegionFocus naturally increases the number of steps taken by the agent. Motivated by this, we investigate whether raising the inference-step limit beyond 100—or removing it altogether—yields further performance gains. Here, inference steps refer to actual browser-interactive actions, excluding RegionFocus overhead. Due to computational constraints, we extended the limit to $300$ steps and also evaluated lower bounds of $10$ and $50$ steps. All the experiments are conducted with the UI-TARS-7B model. As shown in Figure~\ref{fig:ablation_steps}, increasing the maximum to 300 steps improved the 7B model’s performance from $43.2$ to $45.3$, although most trajectories terminated before reaching the 300-step ceiling. Moreover, the incremental benefit gradually decays as the maximum inference-step threshold grows.

\subsection{More Analysis}
\paragraph{Runtime Analysis}
For each trigger, the VLM performs one inference for focal-point proposal, four inferences for region action prediction (which can be executed in parallel), and one inference for action aggregation. The four region action prediction inferences result from adopting four predefined bounding boxes around each predicted focal point, a hyper-parameter. For the interactive environment (WebVoyager), \emph{RegionFocus} occurs in $61.7\%$ of all trajectories; among these, it is triggered an average of $5.84$ times, incurring an average overhead of $66.8\%$ per trajectory. Notably, $32.3\%$ of triggered cases occur only once, with this single trigger yielding an $83.7\%$ improvement in success. For the static environment (ScreenSpot-Pro), \emph{RegionFocus} occurs in $60.2\%$ and $33\%$ of cases for the 72B and 7B models, respectively. The 72B model incurs overheads of $180\%$ when region predictions are executed in parallel, and $360\%$ if executed sequentially. Regarding memory usage, \emph{RegionFocus} requires no additional GPU memory if region predictions are performed sequentially without employing SAM~\cite{kirillov2023segment}. Whether region action predictions should be executed sequentially or in parallel, the number of regions to zoom into, and whether to utilize SAM depend on the specific use case.

\paragraph{Failure Case Analysis}
According to our observations, we summarize the main causes of failures as follows: \emph{(1)}~\emph{Element not visible:} In cases where the correct element is not visible on the current page (e.g., requires scrolling actions), zooming in on the current page is not beneficial. 
\emph{(2)}~\emph{Non-interactable elements:} Some webpages contain extensive text that appears relevant but is not clickable, causing \emph{RegionFocus} to zoom into these regions and subsequently fail to proceed due to the lack of interactable elements. For instance, on the Hugging Face homepage (https://huggingface.co/), there is abundant relevant text such as ``Image-to-Text" and ``Sentence Similarity,” which relates to agent tasks but is non-clickable.
\emph{(3)}~\emph{Wrong focal points:} In some rare cases, \emph{RegionFocus} proposes the wrong focal point, directing attention to irrelevant regions. 
\emph{(4)}~\emph{Action prediction failure:} Occasionally, despite identifying the correct focal point, all regional action predictions fail; our observations suggest that employing SAM [14] could alleviate this issue as it further reduces the background clutter. 
\emph{(5)}~\emph{Action aggregation failure:} Among the regional actions, the selected one may not be optimal.  
\emph{(6)}~\emph{Others:} Other errors arise from general reasoning failures, bot detection, or reaching the maximum step limit.

\section{Conclusion}
\vspace{-0.1in}
We introduced \textit{RegionFocus}, a visual test-time scaling approach that dynamically zooms in on relevant interface regions to address the clutter and ambiguity of modern GUIs. By integrating an \textit{image-as-map} mechanism that marks key landmarks, our method provides transparent action records and improves coordinate-based action predictions. Experiments on Screenspot-pro and WebVoyager show substantial performance gains—even with a simple fixed-ratio bounding box strategy—highlighting the power of visual test-time scaling in enhancing interactive AI systems.

\section{Acknowledgment}
The work was completed during Tiange’s part-time internship at LG AI Research and was supported by an LG AI Research grant awarded through University of Michigan. We thank Jaekyeom Kim and Sungryull Sohn for their discussions and contributions to the codebase.

\clearpage

{
    \small
    \bibliographystyle{ieeenat_fullname}
    \bibliography{main}
}

\clearpage

\appendix

\onecolumn
\section{More experimental details}

In this section, we provide more details about our experimental settings. In our main paper, we examined both the UI-TARS-7B-DPO and UI-TARS-72B-DPO models, as well as Qwen2.5-VL-7B-Instruct and Qwen2.5-VL-72B-Instruct. For WebVoyager, we used a screen resolution of $1440 \times 1440$ pixels for the UI-TARS models and $2240 \times 1260$ for the Qwen models. For both ScreenSpot-Pro and WebVoyager, our predefined bounding boxes were defined as ratios of the input image size, specifically $[0.5, 0.5]$, $[0.3, 0.3]$, $[0.4, 0.8]$, and $[0.8, 0.4]$. Some of the prompts we used are listed below.

\begin{minipage}{1.0\linewidth}
\begin{tcblisting}{
  colback=green!10!white,
  colframe=black,
  listing only,
  width=1.0\textwidth,
  arc=0mm,
  auto outer arc,
  fontupper=\small\ttfamily, %
  title=Prompt for Region Focus,
  listing options={
    breaklines=true,
    basicstyle=\scriptsize\ttfamily, %
    columns=flexible %
  }
}
You are a GUI agent. You are given a task, a current web screenshot, and a history of your previous focused points on the same page (indicated by pink stars in the screenshot). Your job is to output the most relevant point in the screenshot corresponding to the objective. You must avoid the pink-starred coordinates and choose a valid clickable area.

## Other Information
OBJECTIVE: {objective}
URL: {url}

## Output Format
```
(x1, y1)
```
where x1, y1 are the coordinates of the target element, and must differ from any pink-starred coordinates.

## Note
- Ensure the chosen coordinate is a valid clickable area not visibly covered by pink stars in the screenshot.

\end{tcblisting}
\end{minipage}

\begin{minipage}{1.0\linewidth}
\begin{tcblisting}{
  colback=green!10!white,
  colframe=black,
  listing only,
  width=1.0\textwidth,
  arc=0mm,
  auto outer arc,
  fontupper=\small\ttfamily, %
  title=Prompt for Action Prediction -- UI-TARS,
  listing options={
    breaklines=true,
    basicstyle=\scriptsize\ttfamily, %
    columns=flexible %
  }
}
You are a GUI agent. You are given a task and your action history, with screenshots. 
You need to perform the next action to complete the task.

## Other Information
OBJECTIVE: {objective}
URL: {url}

## Output Format
```\nThought: ...
Action: ...\n```

## Action Space
click(start_box='<|box_start|>(x1,y1)<|box_end|>')
left_double(start_box='<|box_start|>(x1,y1)<|box_end|>')
right_single(start_box='<|box_start|>(x1,y1)<|box_end|>')
drag(start_box='<|box_start|>(x1,y1)<|box_end|>', end_box='<|box_start|>(x3,y3)<|box_end|>')
hotkey(key='')
type(content='') #If you want to submit your input, use "\" at the end of `content`.
scroll(start_box='<|box_start|>(x1,y1)<|box_end|>', direction='down or up or right or left')
wait() #Sleep for 5s and take a screenshot to check for any changes.
finished()
call_user() # Submit the task and call the user when the task is unsolvable, or when you need the user's help.

## Note
- Use English in `Thought` part.
- Summarize your next action (with its target element) in one sentence in `Thought` part.
\end{tcblisting}
\end{minipage}

\begin{minipage}{1.0\linewidth}
\begin{tcblisting}{
  colback=green!10!white,
  colframe=black,
  listing only,
  width=1.0\textwidth,
  arc=0mm,
  auto outer arc,
  fontupper=\small\ttfamily, %
  title=Prompt for Action Prediction -- QWen2.5-VL (part 1), 
  listing options={
    breaklines=true,
    basicstyle=\scriptsize\ttfamily, %
    columns=flexible %
  }
}
You are a helpful assistant.

# Tools

You may call one or more functions to assist with the user query.

You are provided with function signatures within <tools></tools> XML tags:
<tools>
{
    "type": "function",
    "function": {
        "name": "computer_use",
        "description": """Use a mouse and keyboard to interact with a computer, and take screenshots.
            * This is an interface to a desktop GUI. You do not have access to a terminal or applications menu. You must click on desktop icons to start applications.
            * Some applications may take time to start or process actions, so you may need to wait and take successive screenshots to see the results of your actions. E.g. if you click on Firefox and a window doesn't open, try wait and taking another screenshot.
            * The screen's resolution is {self.display_width_px}x{self.display_height_px}.
            * Whenever you intend to move the cursor to click on an element like an icon, you should consult a screenshot to determine the coordinates of the element before moving the cursor.
            * If you tried clicking on a program or link but it failed to load, even after waiting, try adjusting your cursor position so that the tip of the cursor visually falls on the element that you want to click.
            * Make sure to click any buttons, links, icons, etc with the cursor tip in the center of the element. Don't click boxes on their edges unless asked."""
        "parameters": {
            "properties": {
                "action": {
                    "description": """
                        The action to perform. The available actions are:
                        * `key`: Performs key down presses on the arguments passed in order, then performs key releases in reverse order.
                        * `type`: Type a string of text on the keyboard.
                        * `mouse_move`: Move the cursor to a specified (x, y) pixel coordinate on the screen.
                        * `left_click`: Click the left mouse button.
                        * `left_click_drag`: Click and drag the cursor to a specified (x, y) pixel coordinate on the screen.
                        * `right_click`: Click the right mouse button.
                        * `middle_click`: Click the middle mouse button.
                        * `double_click`: Double-click the left mouse button.
                        * `scroll`: Performs a scroll of the mouse scroll wheel.
                        * `wait`: Wait specified seconds for the change to happen.
                        * `terminate`: Terminate the current task and report its completion status.
                        """,
                    "enum": [
                        "key",
                        "type",
                        "mouse_move",
                        "left_click",
                        "left_click_drag",
                        "right_click",
                        "middle_click",
                        "double_click",
                        "scroll",
                        "wait",
                        "terminate",
                    ],
                    "type": "string",
                },
                "keys": {
                    "description": "Required only by `action=key`.",
                    "type": "array",
                },
                "text": {
                    "description": "Required only by `action=type`.",
                    "type": "string",
                },
                "coordinate": {
                    "description": "(x, y): The x (pixels from the left edge) and y (pixels from the top edge) coordinates to move the mouse to. Required only by `action=mouse_move` and `action=left_click_drag`.",
                    "type": "array",
                },
\end{tcblisting}
\end{minipage}

\begin{minipage}{1.0\linewidth}
\begin{tcblisting}{
  colback=green!10!white,
  colframe=black,
  listing only,
  width=1.0\textwidth,
  arc=0mm,
  auto outer arc,
  fontupper=\small\ttfamily, %
  title=Prompt for Action Prediction -- QWen2.5-VL (part 2), 
  listing options={
    breaklines=true,
    basicstyle=\scriptsize\ttfamily, %
    columns=flexible %
  }
}
                "pixels": {
                    "description": "The amount of scrolling to perform. Positive values scroll up, negative values scroll down. Required only by `action=scroll`.",
                    "type": "number",
                },
                "time": {
                    "description": "The seconds to wait. Required only by `action=wait`.",
                    "type": "number",
                },
                "status": {
                    "description": "The status of the task. Required only by `action=terminate`.",
                    "type": "string",
                    "enum": ["success", "failure"],
                },
            },
            "required": ["action"],
            "type": "object",
        }
    }
}
For each function call, return a json object with function name and arguments within <tool_call></tool_call> XML tags:
<tool_call>
{"name": <function-name>, "arguments": <args-json-object>}
</tool_call>
\end{tcblisting}
\end{minipage}

\onecolumn

\section{More qualitative results}

\begin{figure*}[ht]
    \centering
    \includegraphics[width=1.0\linewidth]{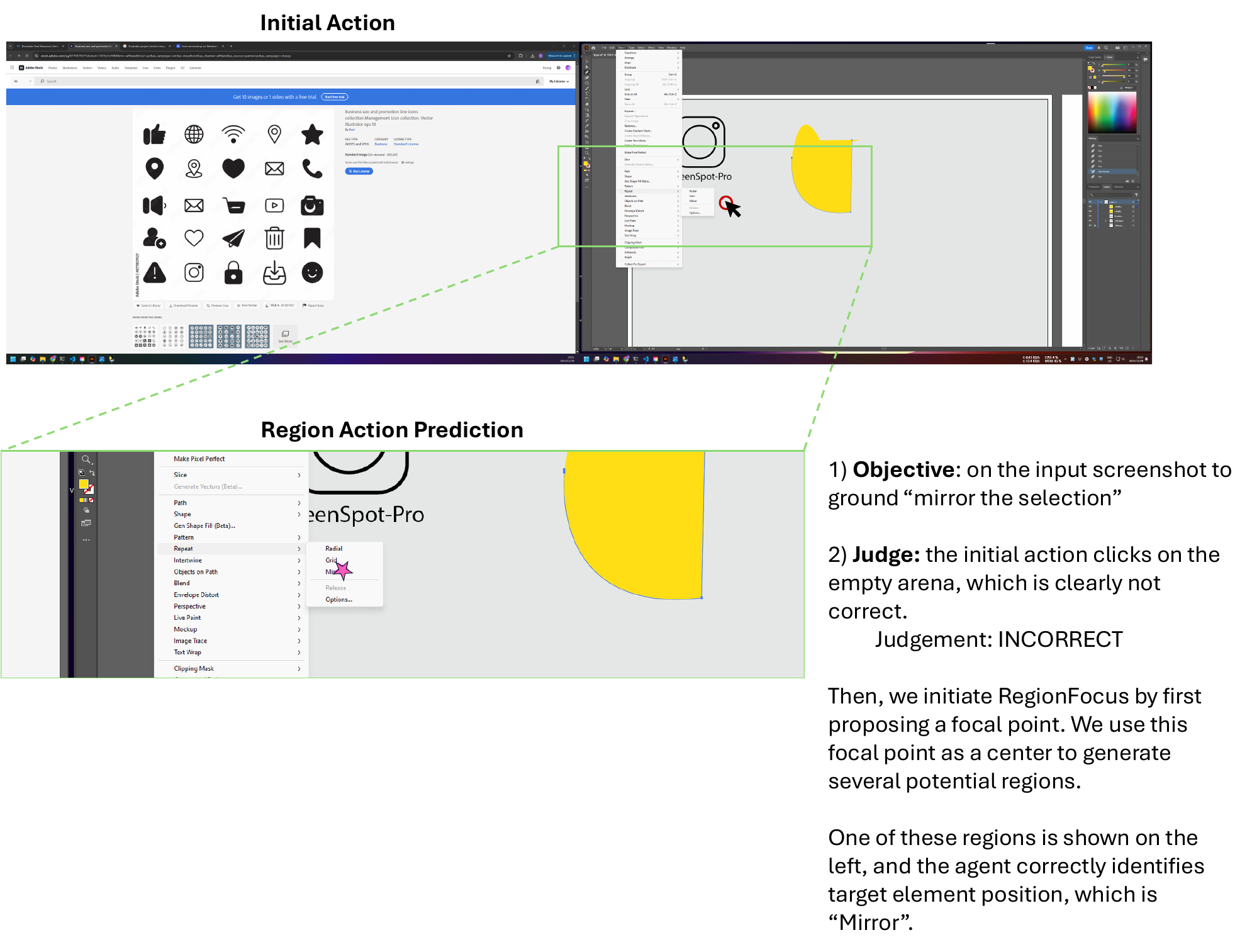}
    \caption{\textbf{Qualitative results - Screenspot-Pro.} In one example from our evaluation, the system successfully rejects the initial action and proposes a correct grounding point based on the zoomed-in view. }
    \label{fig:qual-screenspot3}
\end{figure*}

\begin{figure*}[h]
    \centering
    \includegraphics[width=1\linewidth]{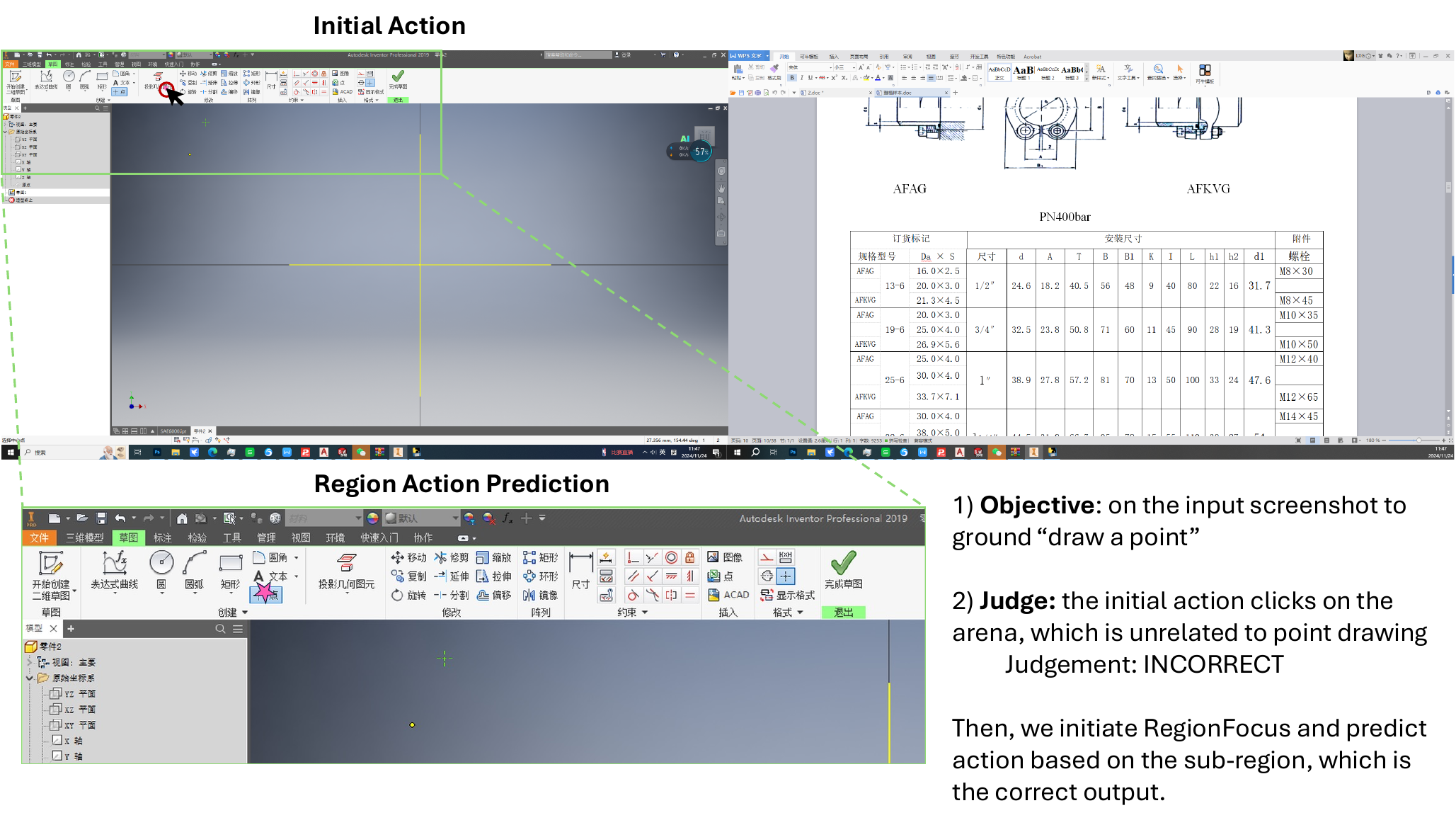}
    \caption{\textbf{Qualitative results - Screenspot-Pro.} In one example from our evaluation, the system successfully rejects the initial action and proposes a correct grounding point based on the zoomed-in view. }
    \label{fig:qual-screenspot2}
\end{figure*}

\begin{figure*}[h]
    \centering
    \includegraphics[width=1\linewidth]{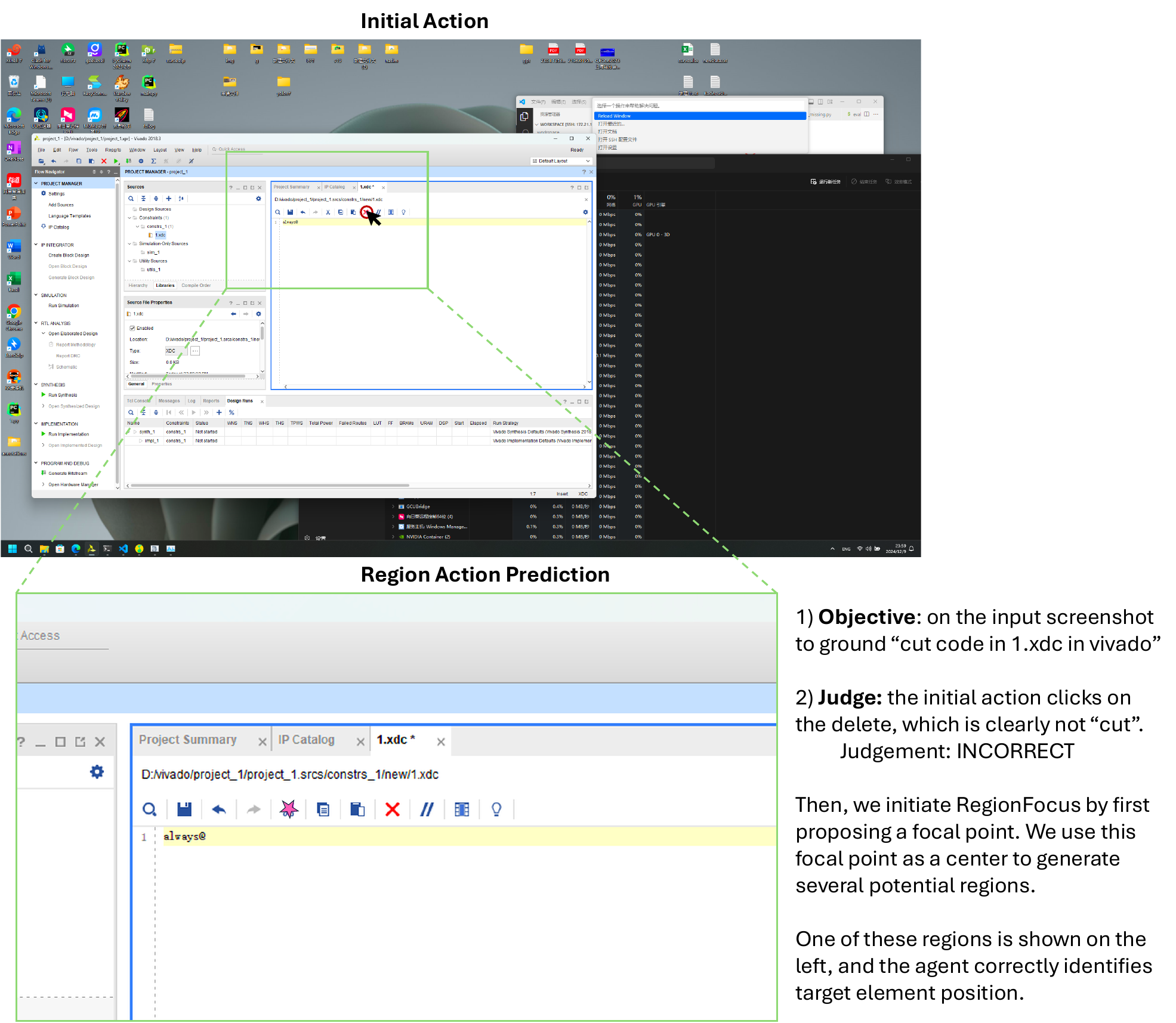}
    \caption{\textbf{Qualitative results - Screenspot-Pro.} In one example from our evaluation, the system successfully rejects the initial action and proposes a correct grounding point based on the zoomed-in view. }
    \label{fig:qual-screenspot4}
\end{figure*}

\end{document}